\documentclass[letterpaper, 10 pt, conference]{ieeeconf}  

\IEEEoverridecommandlockouts                              

\usepackage{xcolor}
\usepackage{commath}

\usepackage{graphics} 
\usepackage{epsfig} 
\usepackage{amsmath} 
\usepackage{amssymb}  
\usepackage{textgreek}
\usepackage{hyperref}
\usepackage{stfloats}
\usepackage{flushend}

\usepackage{caption}
\usepackage{subcaption}

\usepackage{url}
\usepackage{algpseudocode}
\usepackage{algorithm}
\usepackage{comment}
\usepackage{dsfont}
\usepackage{authblk}

\newtheorem{remark}{Remark}

\title{\LARGE \bf
Sim-and-Real Reinforcement Learning for Manipulation:\\ A Consensus-based Approach
}

\author{Wenxing Liu$^{1,3}$, Hanlin Niu$^{1,3}$, Wei Pan$^2$, Guido Herrmann$^3$ and Joaquin Carrasco$^3$
\thanks{This work was supported by EPSRC project No. EP/S03286X/1, EPSRC RAIN project No. EP/R026084/1, EPSRC RNE project No. EP/P01366X/1 and UKAEA/EPSRC Fusion Grant 2022/2027 No. EP/W006839/1.}
\thanks{$^1$Wenxing Liu and Hanlin Niu are with Remote Applications in Challenging Environments (RACE), United Kingdom Atomic Energy Authority, Culham, UK. \textit{(Corresponding author: Hanlin Niu. hanlin.niu@ukaea.uk)}.}
\thanks{$^2$Wei Pan is with the Department of Computer Science, The University of Manchester, Manchester, UK.}
\thanks{$^3$Wenxing Liu, Hanlin Niu, Guido Herrmann and Joaquin Carrasco are with the Department of Electrical \& Electronic Engineering, The University of Manchester, Manchester, UK. }
}

\begin{document}

\maketitle

\thispagestyle{empty}
\pagestyle{empty}

\begin{abstract}
Sim-and-real training is a promising alternative to sim-to-real training for robot manipulations. However, the current sim-and-real training is neither efficient, i.e., slow convergence to the optimal policy, nor effective, i.e., sizeable real-world robot data. Given limited time and hardware budgets, the performance of sim-and-real training is not satisfactory. In this paper, we propose a \underline{C}onsensus-based \underline{S}im-\underline{A}nd-\underline{R}eal deep reinforcement learning algorithm (CSAR) for manipulator pick-and-place tasks, which shows comparable performance in both sim-and-real worlds. In this algorithm, we train the agents in simulators and the real world to get the optimal policies for both sim-and-real worlds. We found two interesting phenomenons: (1) Best policy in simulation is not the best for sim-and-real training. (2) The more simulation agents, the better sim-and-real training. The experimental video is available at: \url{https://youtu.be/mcHJtNIsTEQ}.
\end{abstract}

\section{INTRODUCTION}
As an essential component in robotic control, deep reinforcement learning (DRL) has been widely used in various applications \cite{8968488,han2022deep,han2022reinforcement}. The training process of DRL \cite{sutton1992introduction} builds the bridge between the environment state and the action, thereby maximizing the cumulative reward. Learning from the simulation is safer, cheaper and faster while learning from the real world is more dangerous, expensive and slower. If the simulation shows high fidelity, the training model in the simulation can be transferred directly to the real world. However, in many circumstances, the simulation cannot mimic the real world very well, which limits robot performance in the real world. To overcome this difficulty, we develop a sim-and-real training method to balance the relationship between the simulation and the real world. We use concepts from control engineering, i.e. consensus \cite{wu2021sdp,wu2023finite}, to accomplish sim-and-real training. 

\begin{figure}[!t]
  \begin{center}
    \includegraphics[ width=3.4in]{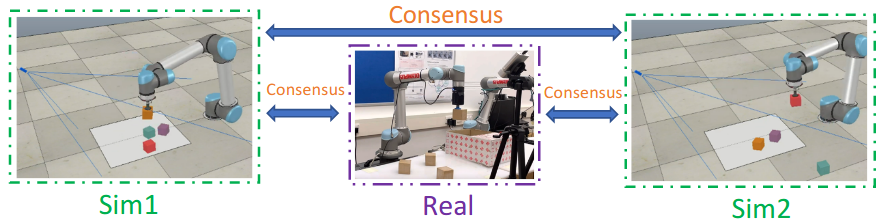}
    \caption{Pick-and-place objects with the CSAR approach}
    \label{fourur5i}
  \end{center}
\end{figure}

In this work, we propose a CSAR algorithm that combines consensus-based training with DRL in a sim-and-real environment, as shown in Fig.~\ref{fourur5i}. We apply CSAR to a group of simulated agents together with a real agent each learning to carry out a pick-and-place task with a suction robot device. Compared to conventional sim-to-real training method, the challenges of CSAR DRL are 1) information exchange between simulated and real robots, for instance, generating communication in a mixed environment, 2) data-efficient collection for training in a sim-and-real environment such as handling data from multiple robots simultaneously, 3) data pre-labelling for suctioning in a sim-and-real environment, for example, using aruco makers to locate suctioned objects. The main contributions of this paper can be concluded as follows:

\begin{enumerate}
\item A complete CSAR method is proposed for manipulators to learn pick-and-place tasks. By applying consensus-based training, the proposed method saves training time and reduces the number of required real robot training steps while maintaining a comparable suction success rate, which is cost-effective.
\item An end-to-end and lightweight neural network is proposed to train the suction policy, which uses raw 3D visual data directly without pre-labelling. The effectiveness and feasibility of the CSAR method are validated through simulation and real-world experiments.  
\item We extend the consensus approach \cite{9833460} from theory and simulations to a real-world pick-and-place problem and show the effectiveness of the proposed approach.
\end{enumerate}


The structure of the paper proceeds as follows. Section II elucidates the related work. Section III details the CSAR algorithm. Experimental validation is given in Section IV to highlight the feasibility of our proposed algorithm. Section V summarizes this paper.

\raggedbottom

\section{RELATED WORK}
\textbf{Sim-to-real:} When a DRL model is transferred from simulation to the real world, the adoption problem becomes challenging as real-world environments contain unpredictable disturbances \cite{peng2018sim}. Fine-tuning has been widely used to bridge the gap between simulated and real environments \cite{kaushik2022safeapt,chen2022bidirectional,bi2021zero}. However, fine-tuning usually takes a long time to perform parameter adaptation, which increases the experimental cost. Some recent works use only simulation but work well in the real world. For instance, with only simulation, a distance function was trained in \cite{viereck2017learning} between the current pose and the nearest optimal pose. In \cite{mahler2017dex}, a grasp quality network was proposed to evaluate robust grasp configuration based on the antipodal grasping sampling method. The key idea of these two papers is to use depth data rather than RGB images since depth images contain less information. Nevertheless, it is challenging for a depth camera to measure thin, dark colour objects because of their physical properties in the real world. Under this condition, performance cannot be guaranteed. Our approach focuses on reducing the gap between simulation and the real world, which is more general and flexible.

\textbf{Sim-and-real:} Sim-and-real training is a recent topic which focuses on adjusting the simulation according to real samples \cite{james2019sim,kang2019generalization}. A novel domain adaptation approach for robot perception was developed in \cite{DBLP:journals/corr/TzengDHFPLSD15} to close the sim-and-real gap by finding common features of real and synthetic data. In \cite{chebotar2019closing}, the agent's parameters in the simulation were updated to match the behaviour in the real world. Compared with \cite{chebotar2019closing}, our approach is capable of dealing with situations that are hard to simulate precisely. In \cite{di2021sim}, one agent was used to select a simulated or real environment with a given probability and to interact. Transitions from all environments were stored in a common replay buffer to update training parameters. In contrast to \cite{di2021sim}, our method creates a sim-and-real environment directly with consensus, which avoids any form or transition but runs simulated and real agents in parallel. This is unique and has not been done before. 
 
\textbf{Reinforcement learning for manipulation:} 
Reinforcement learning has been exploited to deal with robotic tasks \cite{9833460,thuruthel2018model,pradhan2012real}. In our work, we focus on improving training efficiency and saving real-world training costs of sim-and-real DRL for robotic manipulators. An efficient real-time hybrid path planning scheme was proposed in \cite{park2007path} to handle the uncertain dynamics of a robot manipulator by combining the probabilistic roadmap method with DRL. How to modulate the elementary movement of a robot arm through meta-parameters using reinforcement learning was proposed in \cite{kober2011reinforcement}. In \cite{yang2020deep}, a robotic manipulator was trained using DRL to solve the task of grasping an initially invisible object via a sequence of grasping and pushing actions. A high-precision peg-in-hole target task was selected in \cite{inoue2017deep} for force-controlled robotic assembly with DRL. Specifically, the force and moment of the robotic manipulator end effector were chosen as the state. Nevertheless, the authors target solving specific low-level tasks such as motion planning in the work mentioned above. Our method pays attention to high-level tasks by treating each robot manipulator as an agent, which is more general and has a wider range of applications.

\section{METHODOLOGY}

We extend the consensus-based approach in \cite{9833460}, which only focuses on simulations, to sim-and-real scenarios. The proposed effective and efficient CSAR method can increase sim-and-real training speed as well as save real-world training costs with consensus-based training.

\subsection{System Overview}
\begin{figure*}
        \centering
        \includegraphics[width=7in,height=4.8in]{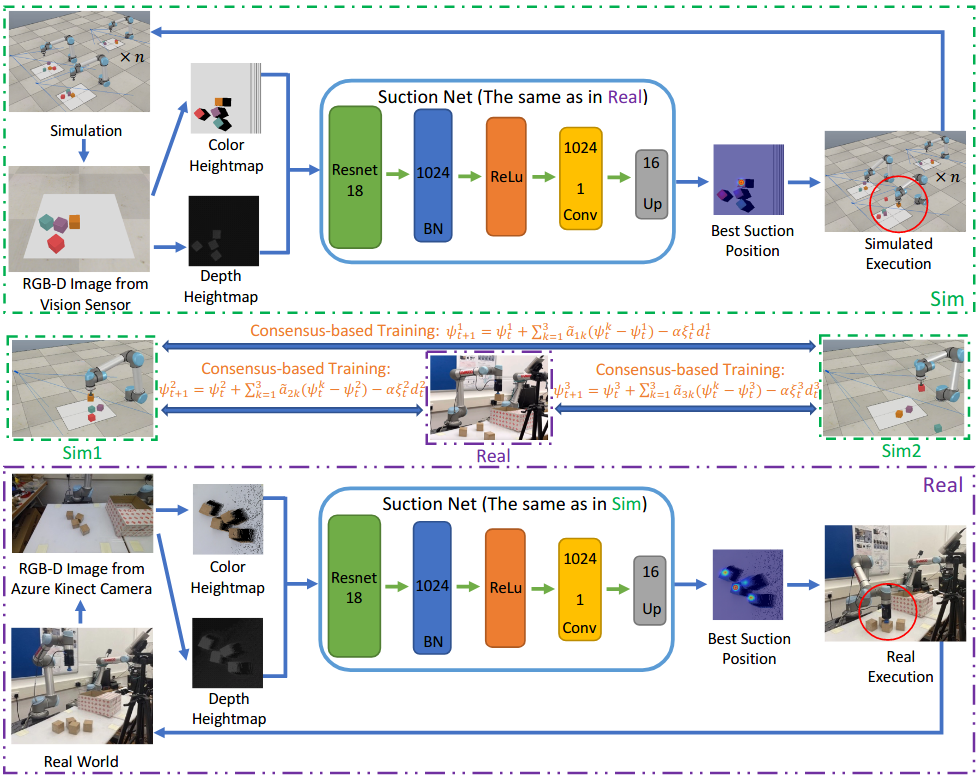}
        \caption{Overview of the proposed DRL framework with consensus-based training in the sim-and-real environment (substantiation of Figure.~\ref{fourur5i}). During each iteration, consensus-based training is applied to the training parameters of every suction net (multi-layer neural network modelling the Q-function for pick-and-place success through suction gripping). The suction executions occur simultaneously in both simulated and real environments. BN represents Batch Normalization. Conv stands for Convolution. Up represents Upsampling. More details can be found in Algorithm \ref{algorithmi}.}

        \label{sim and real}
\end{figure*}

Fig.~\ref{sim and real} describes the overview of our proposed framework. The predefined workspace in the simulation is captured by a fixed simulated camera, which provides an ideal RGB-D image each time. Then the ideal RGB-D image is orthographically projected in the direction of gravity to construct the colour heightmap $\Bar{c}_t$ and the depth heightmap $\Bar{d}_t$, which are the inputs of our framework. Both heightmaps are fed into the Q-function neural network to anticipate pixel-wise best suction position $[\Bar{x}_t, \Bar{y}_t]$. Given the specific use of these neural networks modelling the Q-function for pick and place success through suction gripping, we may call these ``suction networks". The suction height $\Bar{z}_t$ can be found from $\Bar{d}_t$. 

When it comes to the real world, the predefined workspace is captured by a fixed azure kinect camera. Compared with the ideal RGB-D image which is obtained from the simulated camera, the real-world RGB-D image contains more camera distortion \cite{cai2019metagrasp}. Similarly, the real-world RGB-D image is orthographically projected in the direction of gravity to construct the colour heightmap $\Tilde{c}_t$ and the depth heightmap $\Tilde{d}_t$ which are also fed into the suction network to predict real-world pixel-wise best suction position $[\Tilde{x}_t, \Tilde{y}_t]$. The suction height $\Tilde{z}_t$ can be also acquired from $\Tilde{d}_t$.

After performing predictions in both environments, consensus-based training is applied to the training parameters of each simulated or real agent. The suction process of each agent is carried out in parallel, which saves training time.

\subsection{DRL Setup}
\subsubsection{Action Space}
As stated in Section A, the action space $a_t$ is a Cartesian motion command that consists of pixel-wise best suction position. In the simulated environment, $\Bar{a}_t = [\Bar{x}_t, \Bar{y}_t, \Bar{z}_t]$. Correspondingly, $\Tilde{a}_t = [\Tilde{x}_t, \Tilde{y}_t, \Tilde{z}_t]$ in the real world. The suction height $\Bar{z}_t$ and $\Tilde{z}_t$ can be acquired from $\Bar{d}_t$ and $\Tilde{d}_t$. 
\subsubsection{State Space}
As shown in Fig.~\ref{sim and real}, the state space $s_t$ denotes the colour heightmap and depth heightmap of the captured RGB-D image. In the simulated environment, $\Bar{c}_t$ and $\Bar{d}_t$ are acquired by the fixed simulated camera. In the real environment, $\Tilde{c}_t$ and $\Tilde{d}_t$ can be obtained from the fixed azure kinect camera.

\subsubsection{Reward Space}
The distance $\mu_m$ in the simulated environment can be computed by 
\begin{equation} 
\ \mu_m = \sqrt{(\Bar{x}_m-\tau_m)^2+(\Bar{y}_m-\sigma_m)^2}
\end{equation}
where $\tau_m$ and $\sigma_m$ denote $x,y$ positions of the centre of the expected suctioned object of the $m^{th}$ agent, respectively.

We assign suction reward $r_s = 1$ if the target is successfully suctioned, otherwise $r_s = 0$. Thus, the DRL reward $\Bar{r}_m$ for each agent in the simulation can be defined as

\emph{\begin{equation}
\Bar{r}_m = \begin{cases} \label{reward}
r_s r_{0}& \quad \text{if} \,\,\  \mu_m \leq \mu_{th} \\
r_s r_{1} & \quad \text{if} \,\,\  \mu_{th}<\mu_m \leq 2\mu_{th} \\
r_s r_{2} & \quad \text{if} \,\,\  2\mu_{th}<\mu_m \leq 3\mu_{th}\\
r_s r_3 & \quad \text{if} \,\,\  \mu_m >3\mu_{th} 
\end{cases}
\end{equation}}

\noindent where $\Bar{r}_m$ stands for the reward of the $m^{th}$ agent in the simulated environment, $\mu_{th}$ represents distance threshold of the $m^{th}$ agent, $r_{0}, r_{1}, r_{2}$ and $r_{3}$ are the positive reward when $\mu_m$ is within the corresponding range. 

The DRL reward $\Tilde{r}_m$ for each agent in the real environment is given by
\begin{equation}
\Tilde{r}_m = r_s r_{0}
\end{equation}

\begin{algorithm}[!t]
\caption{CSAR: Consensus-based Sim-and-Real DRL} \label{algorithmi}
\begin{algorithmic}[1] 
\State {Initialize the $m^{th}$ agent training parameter $\psi^m_t$, learning rate $\alpha$, RGB-D image $\Bar{g}_t^m$ from the simulation, initial RGB-D image $\Tilde{g}_t^m$ from the real world, discounted factor $\gamma$, total training steps parameter $T$.}

\While {$t < T$} 

\State{Generate $\Bar{c}_t^m$ and $\Bar{d}_t^m$ from $\Bar{g}_t^m$.}
\State{Generate $\Tilde{c}_t^m$ and $\Tilde{d}_t^m$ from $\Tilde{g}_t^m$.}

\If {object count $O_t^m <$ empty threshold} 

\State{Feed $\Bar{c}_t^m$ and $\Bar{d}_t^m$ into the $m^{th}$ suction network to generate action-value function $Q(\psi^m_t,\Bar{s}^m_t,\Bar{a}^m_t)$.}

\State{Feed $\Tilde{c}_t^m$ and $\Tilde{d}_t^m$ into the $m^{th}$ suction network to generate 
action-value function $Q(\psi^m_t,\Tilde{s}^m_t,\Tilde{a}^m_t)$.}

\If {$t > 2$} 

\State{Generate $r^m_t$ with $Q(\psi^m_{t-1},\Bar{s}^m_{t-1},\Bar{a}^m_{t-1})$ and $Q(\psi^m_{t-1},\Tilde{s}^m_{t-1},\Tilde{a}^m_{t-1})$.}

\State{Compute $\xi_{t-1}^m$:\\
$\quad \quad \quad \quad \quad \quad \quad  Y^m_{t-1} = r^m_{t} + \gamma \underset{a}{\max}(Q(\psi^-_{t-1},s^m_{t},a^m))$.\\
$\quad \quad \quad \quad \quad \quad \quad \xi^m_{t-1} = Q(\psi^m_{t-1},s_{t-1}^m,a_{t-1}^m)-Y_{t-1}^m$.}

\State{For $M$ agents, update the training parameters $\psi_{t}$ with consensus-based training:\\
$\quad \quad \quad \quad \quad \quad \quad \quad \psi_{t} =\mathcal{C} \left( \psi_{t-1}, \mathcal{L} \right)-\alpha\Gamma_{t-1}$.}

\State{Sample a batch from the replay buffer $R_p$ to implement experience replay.}

\EndIf

\State{Perform suction execution in both simulated and real environments in parallel.}
\State{Store $(\Bar{c}^m_t, \Bar{d}_t^m, \Bar{a}_t^m)$ and $(\Tilde{c}^m_t, \Tilde{d}_t^m, \Tilde{a}_t^m)$ in $R_p$.}

\Else
\State{Reposition objects.}

\EndIf

\EndWhile

\end{algorithmic} 
\end{algorithm}

\subsubsection{Neural Network Structure}
As stated in Fig.~\ref{sim and real}, the input of the suction net passes data through ResNet-18 \cite{szegedy2017inception} to extract concatenated features from the colour heightmap and the depth heightmap. The aforementioned features are fed into a Batch Normalization layer \cite{paszke2019pytorch} with 1024 input features, a ReLu layer \cite{paszke2019pytorch}, a Convolution layer \cite{paszke2019pytorch} with 1024 input channels, and 1 output channel, then are processed by a bilinear upsample layer \cite{paszke2019pytorch} with a scale factor of 16. The output of the suction net has the same image size as the heightmap input, which is a dense pixel-wise map of different Q values. The pixel which has the maximum Q value represents the best suction position. 
\begin{remark}
It should be noted that the suction net can be substituted by any state-of-the-art neural network. Since we use a standard laptop for training, we purposely design a lightweight version of the suction net inspired by \cite{zeng2018learning}.
\end{remark}

During each training iteration $t$, the training objective is to minimize the temporal difference error $\xi_t$ \cite{mnih2013playing}:
\begin{equation} \label{xi compute}
    \xi_t = Q(\psi_t,s_t,a_t)-{Y_t}
\end{equation}
where $Y_t = r_{t+1} + \gamma \underset{a}{\max}(Q(\psi^-_t,s_{t+1},a))$ and $a$ represents all available actions, $\gamma$ stands for the discount factor, $Q$ represents the action-value function, $r$ is the reward, $\psi_t$ stands for the training parameters of the suction network at time $t$, $\psi^-_t$ denotes the target training parameters.

\subsubsection{Loss function}
Inspired by \cite{zeng2018learning}, we use the Huber loss function \cite{lambert2011robust} to train our proposed suction network in both simulated and real environments. The loss function $\Omega$ at the $t^{th}$ iteration can be computed as follows:
\emph{\begin{equation}
\Omega_t = \begin{cases} 
\frac{1}{2}(\xi_t)^2 & \quad \text{if} \,\,\  |\xi_t|<1\\
|\xi_t|- \frac{1}{2} & \quad \text{otherwise}
\end{cases}
\end{equation}}

Gradients are only passed through the single pixel on which the action is executed during each iteration $t$. All other pixels propagate with 0 loss \cite{zeng2018learning}.

\subsection{Consensus-based Training}
The Q-function of each simulated or real agent is trained through a consensus based algorithm. Hence, we wish to introduce at first the consensus network structure which facilitates that training process. The interaction topology of $M$ agents can be depicted by an undirected graph $\mathcal{G} = (\mathcal{V}, \mathcal{E})$, where $\mathcal{V}$ represents a vertex set $\mathcal{V}=\{1,2,\cdots,M\}$ and $\mathcal{E}$ stands for an edge set $\mathcal{E}\subset\mathcal{V}\times\mathcal{V}$. The edge $(j,m)\in\mathcal{E}$ if the $j^{th}$ and $m^{th}$ agents are connected with one another \cite{wu2022mixed}. The adjacency matrix $\mathcal{A}$ of $\mathcal{G}$ can be described as $\mathcal{A}=[a_{jm}]\in \mathbb{R}^{M\times M}$, where $a_{jm}>0$ if $(j,m)\in\mathcal{E}$, otherwise $a_{jm}=0$. Hence, the Laplacian matrix $\mathcal{L}$ of $\mathcal{G}$ is defined as $\mathcal{L}=\mathcal{D}-\mathcal{A}$, where $\mathcal{D}=diag\{d_{11},\cdots,d_{MM}\}\in\mathbb{R}^{M\times M}$ and $d_{jj}=\sum_{j\neq m}a_{jm}$  \cite{godsil2013algebraic}. For an undirected topology, $\mathcal{L}$ is positive semi-definite. $\mathcal{L}\textbf{1}_{M}= 0$, where $\textbf{1}_{M}=[1,\cdots,1]^{\top}$. If the graph $\mathcal{G}$ has a spanning tree, the rank of $\mathcal{L}$ should be $M-1$ \cite{godsil2013algebraic}. 

For an undirected graph $\mathcal{G}$, if $\hat{\chi}_{m}\in\mathbb{R}^n$ represents the updated training parameter of $\chi_m\in\mathbb{R}^n$ after a single consensus step and $\chi_m$ stands for the row vector of the training parameter for agent $m$ in the graph, the consensus training step of each agent $m$ can be described as
\begin{gather} 
\hat{\chi}_{m} = \chi_m + u_m \label{w1i} \\
u_m = \sum\nolimits_{k=1}^M {a}_{mk}(\chi_k-\chi_m) \label{w2i}
\end{gather}
where ${a}_{mk}$, the element of the graph adjacency matrix, is engendered by the undirected graph $\mathcal{G}$ and $u_m$ stands for the input of the agent $m$. 

By integrating \eqref{w2i} and \eqref{w1i}, the consensus algorithm can be used to update an agent $m$ in the following scheme:
\begin{equation} 
\begin{split} \label{consensusi}
   \hat{\chi}_{m} &=  \chi_m +  \sum\nolimits_{k=1}^M {a}_{mk}(\chi_k-\chi_m)\\
   & =\chi_m-  \sum\nolimits_{k=1}^M{l}_{mk}\chi_k \\
   & =\mathcal{C}_{m}\left(\chi_{k}, l_{mk} \right )
\end{split}
\end{equation}
where ${l}_{mk}$ is the element of the Laplacian matrix $\mathcal{L}$ and $\mathcal{C}_m$ represents the consensus protocol of the $m^{th}$ agent. The training parameter update of all the $M$ agents with a single consensus step can be summarised as
\begin{equation} \label{consensus1i}
\begin{split}
\hat{\chi} & = ((I_{M}-\mathcal{L}) \otimes I_{n}) \chi\\
   &=\mathcal{C} \left( \chi, \mathcal{L} \right)
\end{split}
\end{equation}
where $\mathcal{C}$ stands for the consensus protocol for all agents, $I_{M}$ and $I_n$ denote the $M\times M$ and $n\times n$ identity matrix, $\mathcal{L}$ represents the Laplacian matrix. By repetitively computing \eqref{consensus1i}, this consensus algorithm makes all agents converge to their weighted average \cite{olfati2004consensus}.

\subsection{Consensus-based Training with DRL}
Given the consensus network structure in the previous sub-section, the training algorithm for the training parameters $\psi_t$ in \eqref{xi compute} in the DRL is now introduced. As stated in \cite{mnih2015human}, the process of updating $\psi_t$ for the $m^{th}$ agent is given as:
\begin{equation} \label{lr}
    \psi_{t+1}^m = \psi_t^m - \alpha \xi_t^m \frac{dQ(\psi^m_t,s^m_t,a^m_t)}{d\psi_t^m} 
\end{equation}
where $\alpha$ represents the learning rate.

By applying \eqref{consensusi}, the training process of the CSAR algorithm can be summarised as:
\begin{gather}
\hat{\psi}_t^m = \psi_t^m + \sum\nolimits_{k=1}^M \Tilde{a}_{mk}(\psi_t^k-\psi_t^m) \label{c1}\\
\psi_{t+1}^m = \hat{\psi}_t^m - \alpha \xi_t^m d_t^m \label{c2}
\end{gather}
where $d_t^m = \frac{dQ(\psi^m_t,s^m_t,a^m_t)}{d\psi_t^m}$.

Substituting \eqref{c1} into \eqref{c2}, we can get 
\begin{equation} \label{c3}
      \psi_{t+1}^m = \psi_t^m + \sum\nolimits_{k=1}^M \Tilde{a}_{mk}(\psi_t^k-\psi_t^m)- \alpha \xi_t^m d_t^m 
\end{equation}

Let $\Gamma_t = [ \xi_t^1 d_t^1,  \xi_t^2 d_t^2, \cdots , \xi_t^M d_t^M]^T$, for $M$ agents, the update of the training parameters in our suction network in the $t^{th}$ iteration can be illustrated as
\begin{equation} \label{f1}
\begin{split}
\psi_{t+1} & = ((I_{M}-\mathcal{L}) \otimes I_{n}) \psi_t-\alpha\Gamma_t\\
   &=\mathcal{C} \left( \psi_t, \mathcal{L} \right)-\alpha\Gamma_t
\end{split}
\end{equation}

Algorithm \ref{algorithmi} summarizes our CSAR algorithm.

\section{EXPERIMENTS AND RESULTS}
The feasibility of the CSAR algorithm is validated in this section. The system is implemented on a standard laptop with Nvidia GTX 2070 super and Intel Core i7 CPU (2.6 GHz) with 16 GB RAM. The experimental video is available at: \url{https://youtu.be/mcHJtNIsTEQ}.

\subsection{Experiment Setup}
\subsubsection{Simulation}
Our system in the simulated environment is trained in Coppeliasim \cite{rohmer2013v} with Bullet Physics 2.78 for dynamics, as demonstrated in Fig.~\ref{fourur5i}. The simulation setup for each agent consists of a UR5 robot arm with a suction gripper \cite{ge2021supervised}. The suctioned objects in the simulated environment are cubes with a side length of $5$ cm. The motion planning task for each UR5 robot arm is accomplished by Coppeliasim \cite{rohmer2013v} internal inverse kinematics. Simulated cameras are used to capture RGB-D images of each agent in a $0.448\times0.448$ m$^2$ workspace. The resolution of the simulated RGB-D images is $640\times480$.

\subsubsection{Real World}
The setup for each agent in the real environment is composed of a UR5 robot arm with a Robotiq EPick vacuum gripper. The suctioned objects are cubes with a side length of $6.5$ cm. To pick and place objects successfully with the suction gripper in the sim-and-real environment, the objects should have a flat surface and no overlap between objects placed in the workspace. We use a fixed Azure Kinect camera to acquire real-world RGB-D images with a resolution of $1280\times720$. The location of the Azure Kinect camera is shown in Fig.~\ref{fourur5i}, which can generate a top-down view in a $0.448\times0.448$ m$^2$ workspace. 

\subsubsection{Reward}
Depending on the intrinsic and distortion of the Azure Kinect camera and the size of our suction gripper, we assign $r_{0} = 2000$, $r_{1} = 1000$, $r_{2} = 100$, $r_3 = 1$ and $\mu_{th}=0.005$ m in \eqref{reward}. These values can also be reconfigured for other robotic platforms.

\subsubsection{Neural Network} The proposed framework is fully trained under self-supervision through the interactions between the UR5 robot arms and the sim-and-real environment. The learning rate $\alpha$ in \eqref{lr} has a fixed value of 0.0001. The discounted factor $\gamma$ listed in \eqref{xi compute} is set to 0.5. The future reward discount is fixed at 0.5. The total training steps parameter $T$ is initialized at 270. Algorithm \ref{algorithmi} satisfies $\epsilon$-greedy exploration strategy with $\epsilon$ initialized at 0.5 and annealed to 0.1 over training. The simulated camera and the Azure Kinect camera capture RGB-D images to generate colour and depth heightmaps, which are fed into the suction nets to predict pixel-wise best suction positions.

\subsubsection{Evaluation Metric}
The suction performance of the $m^{th}$ agent can be evaluated using the suction success rate $S^m_r$, which is defined as follows:
\begin{equation}
    S^m_r = \frac{N^m_s}{N^m_i}\times100\%
\end{equation}
where $N^m_s$ represents the number of successful target suctions of the $m^{th}$ agent, $N^m_i$ represents the number of iterations of the $m^{th}$ agent. 

We explore various training strategies to discover the most suitable training conditions for robots:

\textbf{Sim-and-Real:} Only simulation samples are used to train and optimise the model initially. When the suction success rate in the simulation reaches 0.5, we switch to the CSAR method with 3 \textbf{sim}ulated robots and 1 \textbf{real} robot.

\textbf{Sim-to-Real:} Only simulation samples are used to train and optimise the model at the beginning. When the suction success rate in the simulation reaches 0.5, we switch to real-world training with 1 real robot.

\begin{figure}[!ht]
\vspace{-0.3cm}
\centering
\includegraphics[width=2.3in]{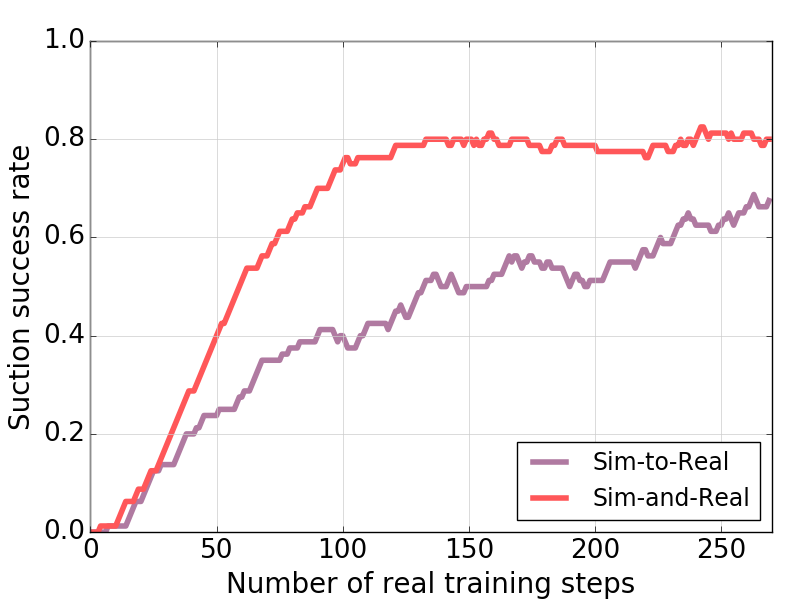}
\caption{Suction success rates of the real robot between “Sim-to-Real” and “Sim-and-Real” strategies}
\label{methodtot}
\end{figure}

\begin{figure}[!ht]
\vspace{-0.5cm}
\centering
\subfloat[]{\includegraphics[width=0.7in]{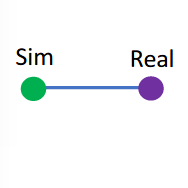}}
\hspace{0.05cm}
\subfloat[]{\includegraphics[width=0.7in]{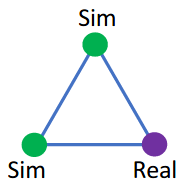}}
\hspace{0.05cm}
\subfloat[]{\includegraphics[width=0.7in]{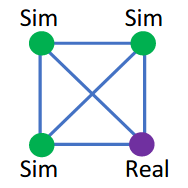}}
\caption{Topology of the interaction of simulation and the real world: (a) 1 \textbf{sim}ulated robot and 1 \textbf{real} robot; (b) 2 \textbf{sim}ulated robots and 1 \textbf{real} robot; (c) 3 \textbf{sim}ulated robots and 1 \textbf{real} robot}

\label{tto}
\end{figure}

\begin{figure}[b]
         \centering
         \vspace{-0.4cm}
         \includegraphics[width=2.3in]{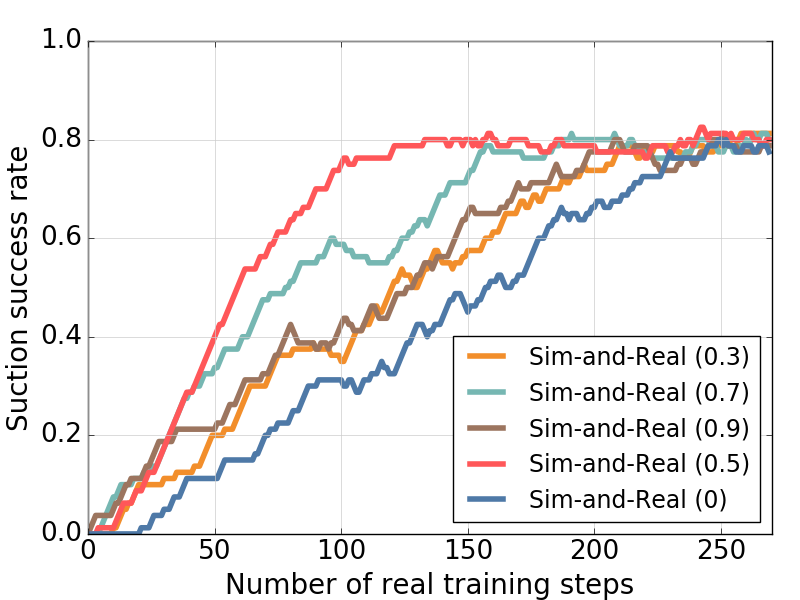}
         \caption{Suction success rates of the real robot with different initial weights when applying the Sim-and-Real strategy. The number in brackets denotes the suction success rate from the pre-trained simulation model.}

         \label{meth}
     \end{figure}
\vspace{-0.4cm}
\subsection{Sim-and-Real is Better Than Sim-to-Real}

Fig.~\ref{methodtot} demonstrates the suction success rate of the real robot using two different training strategies. The interaction topology of Sim-and-Real is shown in Fig.~\ref{tto} (c). When applying the Sim-and-Real strategy, the suction success rate of the real robot reaches 80\% at around 140 training steps, which outperforms the Sim-to-Real strategy. Since our policy for each robot is greedy deterministic, a robot may execute the same action repetitively if there is no environment change when using the Sim-to-Real training strategy. However, by applying consensus-based training, the simulated agent can be used to introduce noise indirectly into the sim-and-real environment, which prevents robots from getting stuck in the same action. In summary, applying the Sim-and-Real strategy leads to a faster training speed, which saves real-world training costs.

\subsection{Best Policy in Simulation is Not the Best for Sim-and-Real Training}

A striking observation from our experiment is that the best-obtained policy trained in simulation is not the best pre-trained model to start the co-training between simulated and real robots, as shown Fig.~\ref{meth}. When the suction success rate of the pre-trained simulation model is 0.5, the Sim-and-Real strategy achieves the best performance. When the suction success rate drops to 0.3, it takes longer for the real robot to solve the task. Surprisingly, when the suction success rate of the pre-trained simulation model is too high (0.7, 0.9), the performance deteriorates.

This is counterintuitive, as shown in the Sim-to-Real experiment, that the best policy obtained in the simulation is typically the one to be deployed. This observation suggests that the ``mediocre" policy is the best for co-training. When the success rate of the pre-trained simulation model is too high, the sim-and-real framework will be initialised at a value that is close to the optimal simulation value. This will take longer to converge to the mixed optimality in a sim-and-real environment. As a result, applying the ``mediocre" policy can reduce real robot training costs and save the pre-training time in the simulation.

\subsection{The More Agents in Simulation, the Better for Sim-and-Real Training}
\vspace{-0.3cm}
\begin{figure}[!ht]
\centering
\includegraphics[width=2.3in]{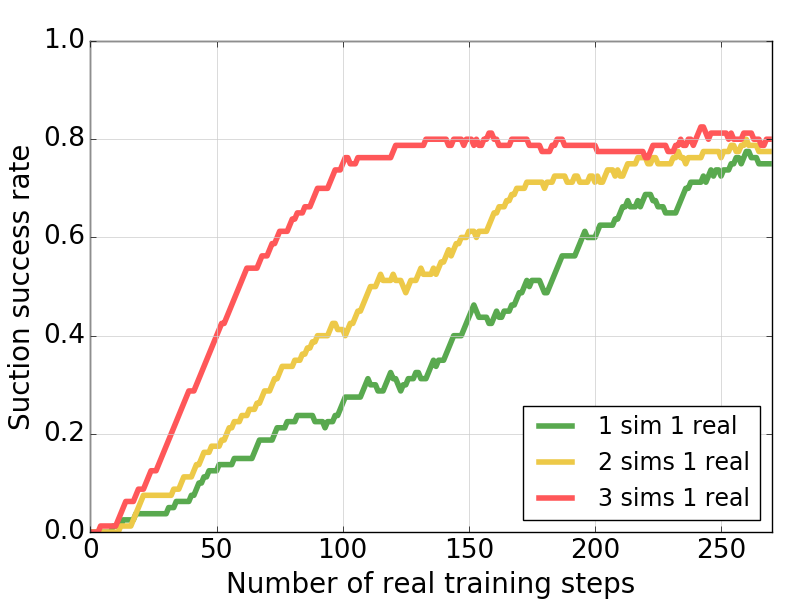}
\caption{Suction success rates of the \textbf{real} robot with different number of \textbf{sim}ulated robots using Sim-and-Real strategy}
\label{diff}
\end{figure}

Readers may wonder why we use 3 \textbf{sim}ulated robots and 1 \textbf{real} robot during training. Therefore, we vary the number of simulated robots when using the Sim-and-Real strategy. Fig.~\ref{diff} describes the suction success rate when using the Sim-and-Real strategy with different numbers of simulated robots. The interaction topology used in Fig.~\ref{diff} is shown in Fig.~\ref{tto}. It takes around 260 steps to make the real robot arrive at 80\% suction success rate when using 1 \textbf{sim}ulated robot and 1 \textbf{real} robot strategy. In the case of 2 \textbf{sim}ulated robots 1 \textbf{real} robot, the required training steps descend to around 240. Only around 140 steps are required to maintain the same suction success rate when using the 3 \textbf{sim}ulated robots 1 \textbf{real} robot strategy. More simulated robots participating in the proposed framework can accelerate the training speed and exhibit good robustness in the sim-and-real environment, thus decreasing the number of required real robot training steps while maintaining a comparable suction success rate.

\subsection{Generalisation of Real-world Unseen Objects}
\begin{figure}[!ht]
\centering
\subfloat[]{\includegraphics[width=0.9in]{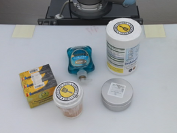}}
\hspace{0.05cm}
\subfloat[]{\includegraphics[width=0.9in]{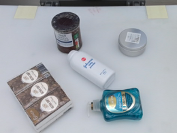}}
\hspace{0.05cm}
\subfloat[]{\includegraphics[width=0.9in]{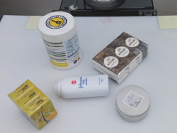}}
    \caption{Novel objects for validation: (a) Environment 1; (b) Environment 2; (c) Environment 3}
    \label{g all}
\end{figure}

\begin{figure}[!ht]
  \vspace{-0.4cm}
  \begin{center}
   \includegraphics[width=3.3in]{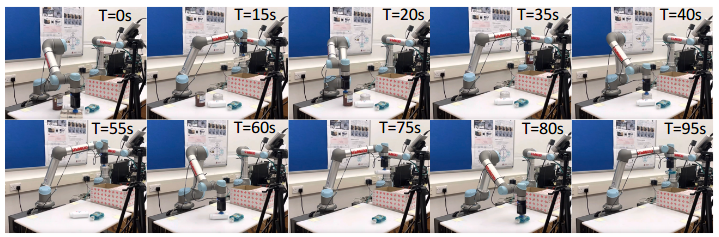}
    \caption{The demonstration of picking novel objects. More details can be seen in the video.}
    \label{gdemoreal}
  \end{center}
\end{figure}

The Sim-and-Real strategy is capable of generalising to novel objects (Fig.~\ref{g all}) with a suction success rate of $80\%$. After training on cubes in both simulated and real environments, the CSAR training model can also be applied to pick and place novel objects such as cylinders and irregularly shaped objects with different heights, as shown in Fig.~\ref{gdemoreal}.

\section{CONCLUSION}
In this work, we propose a CSAR approach which is able to improve sim-and-real training speed and reduce real-world training costs. By implementing the Sim-and-Real strategy, the suction success rate of the real robot attains 80\% at around 140 training steps, which outperforms the Sim-to-Real strategy. Applying the ``mediocre" policy can not only reduce the number of required real robot training steps but also save the pre-training time in the simulation. More simulated robots participating in the CSAR method increase the training speed, thereby reducing real-world training expenses. The Sim-and-Real strategy is also capable of generalising to novel objects. The CSAR method is a straightforward generalization and practical verification of the team's recently developed theory of a consensus-based RL approach \cite{9833460}. In the future, an optimisation of the CSAR approach will be exploited to tackle more complicated scenarios.






\bibliographystyle{IEEEtran}
\bibliography{bibtex/bib/ref.bib}

\end{document}